\documentclass[12pt]{article}
\usepackage{fullpage}
\usepackage{times}
\newcommand{\omt}[1]{}
\begin{document}

\title{A Matter of Opinion:  Sentiment Analysis and Business
Intelligence \\ (position paper)}
                               
\author{Lillian Lee,  Cornell University \\ http://www.cs.cornell.edu/home/llee}
\date{}

\begin{center}
{\Large A Matter of Opinion:  Sentiment Analysis and Business
Intelligence (position paper)} \\
{\Large Lillian Lee,  Cornell University} \\

\medskip

Presented at the IBM Faculty Summit on the Architecture of On-Demand 
Business, May 2004
\end{center}

\medskip

\paragraph{Motivation}

In the novel  {\em Hard Times}, Charles Dickens described the fictional
``Coketown'' as follows:
\begin{quotation}
Fact, fact, fact, everywhere in the material aspect of the town; fact,
fact, fact, everywhere in the immaterial. The M'Choakumchild
school was all fact, and the school of design was all fact, and the
relations between master and man were all fact, and {everything was
fact between the lying-in hospital and the cemetery, and what you
couldn't state in figures, or show to be purchasable in the
cheapest market and salable in the dearest, was not}, and never should
be, world without end, Amen.
\end{quotation}
In 
real-life business intelligence,  
facts are of course very important, but {\em opinion} also plays a
crucial role.  Consider, for instance, the following scenario.  
A major computer manufacturer, disappointed with unexpectedly low
sales,  finds itself confronted with this question:
\begin{center}
Why aren't consumers buying our laptop?
\end{center}
While concrete data such as the laptop's weight or the
price of a competitor's model are obviously relevant, answering this
question requires focusing more on people's personal {\em views}
of such objective characteristics. Moreover, 
subjective judgments regarding intangible qualities --- e.g., ``the
design is tacky'' or ``customer service was condescending'' --- or
even misperceptions --- 
``updated device drivers aren't  available''
--- 
must be taken into account as well.

{\em Sentiment-analysis technologies} for extracting opinions from
unstructured human-authored documents would be excellent tools for
handling many business-intelligence tasks related to the one just
described.  Continuing with our example scenario: it would be
difficult to try to directly survey laptop purchasers who {\em
haven't} bought the company's product.  Rather, we could employ a system that
(a) finds reviews or other expressions of opinion on the Web ---
newsgroups, individual
blogs, and  aggregation sites such as epinions.com are likely to be productive sources --- and then (b) creates
condensed versions of the reviews or a digest of the overall
consensus. This would save
the analyst from having to read potentially dozens or even hundreds of
versions of the same complaints.  Note that Internet sources can vary
wildly in form, tenor, and even grammaticality; this fact underscores
the need for robust techniques even when only one language (e.g.,
English) is considered.

\paragraph{Challenges in sentiment classification}

Given the multitude of potential applications, 
researchers have been devoting more and more
attention to sentiment analysis.  Much of the current work is devoted
to {\em classification} problems: determining whether a particular
document or portion thereof is subjective or not, and/or determining
whether the opinion it expresses is positive or negative.  At first
blush, this might not appear so hard: one might expect that we need
simply look for obvious sentiment-indicating words, such as ``great''.
The
difficulty lies in the richness
of human language use.  First, there can be
an amazingly large number of ways to say the same thing (especially,
it seems, when that thing is a negative perception); this complicates
the task of finding a high-coverage set of indicators.
Furthermore, the
same indicator may admit several different interpretations.  
Consider, for example,  the following sentences:
\begin{itemize}
\item This laptop is \underline{a great deal}.  
\item \underline{A great deal} of media attention surrounded the release of the new
  laptop model.
\item If you think this laptop is \underline{a great deal},
I've got a nice bridge you might be interested in.
\end{itemize}
Each of these sentences contains the three words ``a great deal'', but the opinions  expressed
are, respectively, positive, neutral, and negative.  The first two
sentences use the same phrase to mean different things.
The last sentence involves sarcasm, which, along with related rhetorical
devices, is an intrinsic feature of texts from unrestricted domains such as
blogs and newsgroup postings.  

In general, researchers have adopted one of two approaches to
meeting the challenges that sentiment analysis presents.
Many groups are working to directly improve
the selection and interpretation of indicators through the
incorporation of linguistic knowledge; given  the
subtleties of natural language, such efforts will be critical to building
operational systems.  
Others have been pursuing a different tack: 
employing  {\em learning
algorithms} that can automatically infer from text samples what
indicators are useful.  Besides being potentially more cost-effective, more
easily ported to other domains and languages, and more robust to
grammatical mistakes, learning-based systems can also
discover indicators that humans might neglect.  For example, in our
own work, we
found that the phrase ``still,'' (comma included)
is a better indicator of positive sentiment than ``good'' --- a
typical instance of use would be a sentence like ``Still, despite these flaws, I'd go with this laptop''. 
Nevertheless,  it bears repeating that incorporating deep knowledge about language
will be absolutely crucial to  developing systems capable of
high-quality (as opposed to merely high-throughput) sentiment analysis.
Both the linguistic and the learning approach have considerable
merits; 
it seems very safe to say that  the community will need to  turn towards
finding ways  to combine their advantages.

\paragraph{Related problems, new directions}

The classification problems discussed above only involve the
determination of sentiment.  However, there is growing interest in
capturing interactions between {\em subjectivity} and
{\em subject} --- we not only need to know what an author's opinion
is, but what that opinion is about.  For example, while in a broad
sense a review of a particular laptop is only about one topic
(the laptop itself), it almost surely discusses various specific
aspects of
the machine.
We would ideally like a sentiment-analysis system to reveal whether
there are particular features that the review's author disapproves of
even if his or her overall impression was positive.

Another interesting research direction of potentially great importance
is to integrate into sentiment analysis the notion of the {\em status}
of an opinion holder, perhaps via  adaptation of the
hubs-and-authorities techniques used in Web search or link-analysis
methods in reputation systems.  For example,
we might want to identify {\em bellwethers} --- thought leaders
with enough influence that others explicitly adopt their opinions ---
or  {\em barometers} --- those whose opinions
are generally held by the majority of the population of
interest. Tracking the views of these two types of people could both
streamline and enhance the process of gathering business intelligence
to a large degree.  Surely that sounds like a great deal!

\omt{Cornell (in general.  And, Claire; Thorsten = textcat)}

\end{document}